\documentclass[journal]{IEEEtran}
\ifCLASSINFOpdf
\else
\fi
\usepackage{amsmath,epsfig}
\usepackage{amsfonts}
\usepackage{multirow}
\usepackage{bm}
\usepackage{cite}
\usepackage{array,color}
\usepackage{stfloats}
\usepackage{balance}
\usepackage{multirow}
\usepackage{url}
\usepackage[colorlinks, linkcolor=red, anchorcolor=blue, citecolor=green]{hyperref}
\begin{document}

\title{Incorporating Lambertian Priors into Surface Normals Measurement }

\author{Yakun Ju,
Muwei Jian,
Shaoxiang Guo,
Yingyu Wang,
Huiyu Zhou,
	    Junyu Dong

\thanks{Y. Ju, S. Guo, Y. Wang, and J. Dong are with the Department of Computer Science and Technology, Ocean University of China, Qingdao, China (e-mails: \{juyakun, guoshaoxiang, wangyingyu\}@stu.ouc.edu.cn, dongjunyu@ouc.edu.cn.)} %
\thanks{M. Jian is with the School of Information Science and Engineering, Linyi University, Linyi, China, and the School of Computer
Science and Technology, Shandong University
of Finance and Economics, Jinan, China (e-mail: jianmuweihk@163.com).} 
\thanks{H. Zhou is with the Department of Informatics, University of Leicester, Leicester, UK (e-mail: hz143@leicester.ac.uk).} 

\vspace*{-5mm}
}

\maketitle

\begin{abstract}
The goal of photometric stereo is to measure the precise surface normal of a 3D object from observations with various shading cues. However, non-Lambertian surfaces influence the measurement accuracy due to irregular shading cues. Despite deep neural networks have been employed to simulate the performance of non-Lambertian surfaces, the error in specularities, shadows, and crinkle regions is hard to be reduced. In order to address this challenge, we here propose a photometric stereo network that incorporates Lambertian priors to better measure the surface normal. In this paper, we use the initial normal under the Lambertian assumption as the prior information to refine the normal measurement, instead of solely applying the observed shading cues to deriving the surface normal. Our method utilizes the Lambertian information to reparameterize the network weights and the powerful fitting ability of deep neural networks to correct these errors caused by general reflectance properties. Our explorations include: the Lambertian priors  (1) reduce the learning hypothesis space, making our method learn the mapping in the same surface normal space and improving the accuracy of learning, and (2)  provides the differential features learning, improving the surfaces reconstruction of details. Extensive experiments verify the effectiveness of the proposed Lambertian prior photometric stereo network in accurate surface normal measurement, on the challenging benchmark dataset.
\end{abstract}

\begin{IEEEkeywords}
Photometric stereo, surface normal measurement, prior fusion, non-Lambertian, deep neural networks
\end{IEEEkeywords}

%
\IEEEpeerreviewmaketitle

\section{Introduction}

\IEEEPARstart{M}{easurement} of 3D geometry from 2D scenes is a key problem in machine vision and industrial applications \cite{ren2017curve,ren2021complex,ju2020dual,jian2019learning}. Unlike multi-view stereo and binocular that use different viewpoints scenes to triangulate sparse 3D points, photometric stereo \cite{Woodham1980Photometric} measures the pixel-wise surface normal from a fixed scene under varying shading cues. The photometric methods prevail in measuring fine details of the surface and dense reconstruction, while it bases on the Lambertian assumption, hardly handling the general reflectance properties existing in real-world objects and deviating from realistic applications. To deal with the limitations, previous research adopted the bidirectional reflectance distribution functions (BRDFs) to model general reflectance \cite{alldrin2008photometric,ikehata2014photometric,higo2010consensus} or treated the non-Lambertian regions as anomalies \cite{wu2010robust,ikehata2012robust,miyazaki2010median}. However, these traditional methods are accurate for a limited class of surface reflectance and suffer from unstable optimization.

\begin{figure*} [!htp]
  \includegraphics[width=1\textwidth]{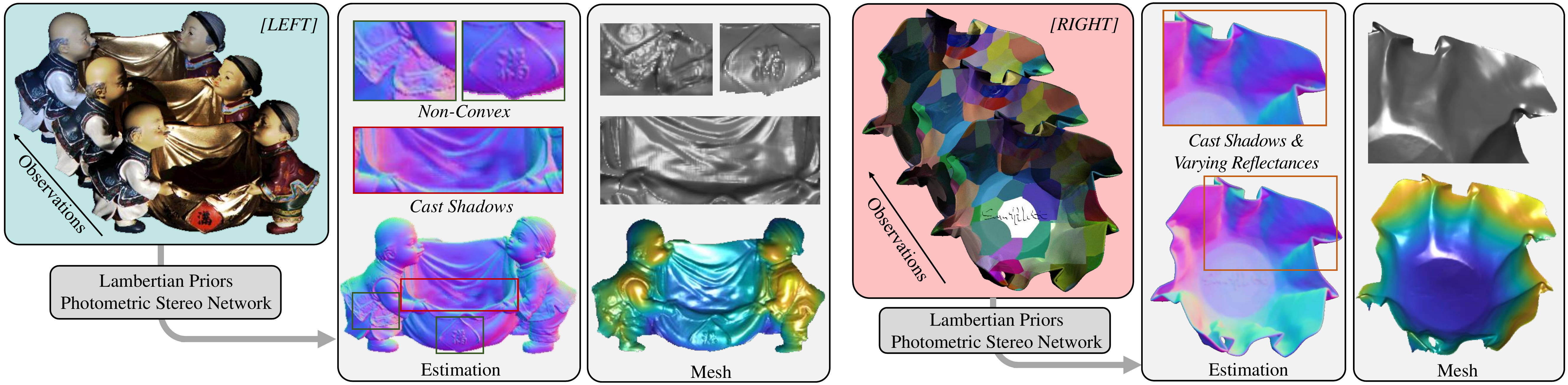}
  \caption{We present a novel Lambertian prior guided photometric stereo network to measure the surface normal of the target. Our method improves the accuracy of strong non-Lambertian surfaces, non-convex structure, and varying reflectance surface. \textit{[LEFT]}: the results on ``Harvest'' of the DiLiGenT benchmark dataset \cite{shi2019benchmark} (using 96 observations), where the strong cast shadows and non-convex structures exist in the observations. \textit{[RIGHT]}: the results on ``Paperbowl (Specular)'' of the CyclesPSTest dataset \cite{ikehata2018cnn} (using 17 observations). The cast shadows and varying reflectance are displayed in the observation. The details of 3D mesh are also shown in each example, which is derived from the measured surface normal by the standard integration method \cite{simchony1990direct}.}
  \label{fig1}
\end{figure*}

Recently, deep learning-based methods have been introduced to photometric stereo \cite{santo2017deep,chen2018ps}. These methods directly learn the surface normal of objects from the observed images, where the capability of handling diverse BRDFs has been proved to be notable. However, strong non-Lambertian and varying reflectance regions still lead to significant errors, in both the traditional algorithms and deep learning-based methods. We reckon that the failure in these regions is due to the fact that (1) the regions and surfaces rarely appear in the training dataset, (2) the overexposed values in specularity and the dark areas in shadows are difficult to produce useful features, (3) the surface with spatially-varying reflectance causes measurement errors on the normal map. These challenging problems remain to be solved. We show our state-of-the-art performance over these conditions in Fig. \ref{fig1}.

To address these challenges, we propose a Lambertian model guided photometric stereo network to better handle non-Lambertian surfaces. Instead of directly embedding the observed shading cues on the surface normal, we propose a strategy that utilizes the observations to modify the Lambertian priors, in other words, our method focuses on the use of the input images as the patterns to correct the measurement of the surface normal from the Lambertian priors. Compared with previous methods mainly focusing on network architectures, our method has the following advantages: First, our method utilizes a mapping in the same surface normal space $\{ \mathcal{Y} \to \mathcal{Y} \}$,  where the learning space is reduced, while the previous methods take a mapping from the RGB image space to the surface normal space $\{ \mathcal{X} \to \mathcal{Y} \}$.  Second, the Lambertian priors and ground-truth surface normal are theoretically similar in terms of diffuse reflection. In this condition, our method is equivalent to enlarge the proportion of non-Lambertian errors in total errors, where the network always tends to learn the parameters to make the loss drop. In training, the network therefore will be more inclined to optimize this part of errors. \emph{i.e.}, our method learns the differential features, amplifying the non-Lambertian errors, where the details of measurement is improved. In short, our method reduces the learnable hypothesis space and improves the estimation results. The experimental results have demonstrated the effectiveness of the Lambertian priors: Our architecture notably improves the accuracy of surface normal measurement, outperforming several state-of-the-art calibrated photometric stereo approaches on the widely used DiLiGenT benchmark dataset \cite{shi2019benchmark}. Furthermore, the ablation experiments show the fast convergence and more accurate surface normal measurement by the proposed photometric stereo network.

The main two contributions of this work are summarized as follows:
\begin{itemize}
\item We propose a Lambertian prior guided photometric stereo network to better handle non-Lambertian surfaces. Our method reduces the learnable hypothesis space and improves the surface normals measurement.
\item To the best of our knowledge, this is the first work that involves embedding the priors into the learning-based photometric stereo network. This strategy can be also used to refine wider non-ideal photometric stereo measurements, by using inaccurate priors. 
\end{itemize}

\section{Related Work}

A measurement pixel of a real-world object $I_j$ can be modeled by general BRDFs $\rho$, which is associated with the surface normal $\boldsymbol{n} \in \mathbb{R}^3$, illumination direction $\boldsymbol{l_j} \in \mathbb{R}^3$, and view direction $\boldsymbol{v} \in \mathbb{R}^3$. With the illumination intensity $e$, the imaging model can be expressed as:
\begin{equation}
\label{imaging}
I_j= e \rho\left(  \boldsymbol{{n}}, \boldsymbol{l_j},\boldsymbol{v} \right) \max \left(\boldsymbol{\bar{n}}^{\top} \boldsymbol{l_j}, 0\right)+\epsilon_j,
\end{equation}
where $\max \left(\boldsymbol{n}^{\top}\boldsymbol{l}_{j}, 0\right)$ revealed the attached shadows, and $\epsilon_j$ is an error term where the impacts are hardly represented by the BRDF model, such as cast shadows, imaging noise, and inter-reflections \cite{nayar1991shape}. As an inverse problem, the goal of photometric stereo is to measure the surface orientation from a combination of reflectance and illuminations in multiple images. The literature of photometric stereo is vast, and we here briefly review the mainstream calibrated photometric stereo technologies, including traditional algorithms and deep learning based methods. Comprehensive photometric stereo surveys on hand-craft reflectance models and robust methods can be found in \cite{ackermann2015survey,herbort2011introduction}.

\subsection{Lambertian Photometric Stereo}
In 1980, Woodham firstly proposed the Lambertian photometric stereo algorithm \cite{Woodham1980Photometric}. Under the Lambertian assumption, the error term $\epsilon_j$ is ignored and the BRDF has the diffuse property, where the observed intensity is proportional to the cosine of the angle between the illumination direction and the surface normal but irrelevant to the view direction. Therefore, the imaging model can be simplified and easily solved by the least-square approach. Although the Lambertian method failed to estimate most of the real-world non-Lambertian objects, the basic theory is significant that reveals the image formation model can be cast into a linear system of equations and solved, establishing the relationship between two-dimensional images and the object geometry.

\subsection{Non-Lambertian Photometric Stereo}
To extend photometric stereo to work with non-Lambertian surfaces in practice, researchers investigated different strategies. Commonly, according to the taxonomy of \cite{shi2019benchmark}, non-Lambertian photometric stereo technologies can be divided into four categories: robust methods, analytic and empirical reflectance model methods, example-based methods, and deep learning methods.

\subsubsection{Robust methods}
Robust methods treat most regions on the surface as a simple diffuse reflectance model (Lambertian) while treating non-Lambertian phenomena (such as specularity and cast shadows) as outliers. These methods assume specularity and shadows are local and sparse, which can be detected and discarded. In the early work, the surface normal is estimated by selecting the three images with the lowest specularity and the closest Lambertian appearance from multiple images \cite{solomon1996extracting}. Afterwards, Wu \textit{et al.} \cite{wu2010robust} proposed a robust principal component analysis (RPCA) method to decompose images into the minimized-rank Lambertian composition and the non-Lambertian sparse counterpart. Similarly, Ikehata \textit{et al.} \cite{ikehata2012robust} employed an improved rank = 3 decomposition instead of rank-minimization, achieving better computational stability. Several other robust techniques were also applied to solve the outliers, such as maximum-likelihood estimation \cite{verbiest2008photometric}, shadow cuts \cite{chandraker2007shadowcuts}, and maximum feasible subsystem \cite{yu2010photometric}. Although robust methods are effective, these approaches generally cannot handle the surface with broad and soft specularity, where non-Lambertian regions are hard to be detected as the outliers. In addition, these methods usually need a huge number of the observed images to achieve stable computations.

\subsubsection{Analytic and Empirical Reflectance Model Methods}
To handle the non-Lambertian, using analytic or empirical reflectance model to approximate the non-Lambertian BRDFs is a fairly straightforward idea. Along this direction, many models were proposed to fit the nonlinear analytic BRDF, such as the specular spike model \cite{chen2006mesostructure}, the Blinn-Phong model \cite{tozza2016direct}, the Torrance-Sparrow model \cite{georghiades2003incorporating}, the Ward model and its variations \cite{chung2008efficient,goldman2009shape}, and the microfacet BRDF model \cite{chen2017microfacet}. In addition, empirical reflectance models consider the general properties of a BRDF, such as isotropy and monotonicity, to deal with multiple types of surface materials. Some basic derivations for isotropy BRDFs were proposed in \cite{chandraker2012differential,alldrin2007toward}. Based on empirical models, some researchers applied isotropic constraints for the measurement of surface orientation \cite{li2015photometric,shi2012elevation}. Shi \textit{et al.} \cite{shi2014bi} and Ikehata \textit{et al.} \cite{ikehata2014photometric} further approximated the isotropic BRDFs by bivariate functions to deal with the instability of the estimation. However, these hand-crafted analytic and empirical reflectance models are generally useful for limited categories of reflectance as the reflectance properties are significantly changing from materials to materials. Moreover, most of these methods are pixel manners which ignore inter-reflection and cast shadows.

\subsubsection{Example Based Methods}

In addition to the above two strategies, amounts of photometric stereo algorithms can be treated as example based methods. Under the same illumination environment, the calibration object with the known surface normal (usually a sphere) transformed the non-Lambertian photometric stereo to a pixel matching problem. Early work required exactly the same material between the target and the calibrated objects. Hertzmann and Seitz \cite{hertzmann2005example} relaxed this restriction by assuming that a small number of basis materials compose the general materials. Hui and Sankaranarayanan \cite{hui2016shape} used a BRDF dictionary to render virtual spheres instead of putting the real calibrated objects. However, the drawback of the same illumination configuration limits its practical use.

\subsubsection{Deep Learning Methods}

Deep learning techniques have been introduced to solve the problem of non-Lambertian photometric stereo, which have achieved inspiring performance. Santo \textit{et al.} first presented DPSN \cite{santo2017deep} to address the non-Lambertian photometric stereo. DPSN regresses the per-pixel normal from the fixed number of the observed images, where the training and the testing have to use the same pre-defined illumination directions. Therefore, a new model has to be retrained if the input number or illumination directions are changed. To relax this constraint and take advantage of the information embedded in the neighborhood, subsequent methods were improved by applying convolutional neural networks (CNNs) and explored flexible input strategies \cite{ikehata2018cnn,li2019learning,chen2018ps}. Ikehata \cite{ikehata2018cnn} introduced the observation map, which rearranges observation intensities according to light directions, to overcome the fixed inputs problem. The observation map strategy was also adopted in \cite{li2019learning,zheng2019spline}, which is effective for inputs with order-agnostic illuminations. Others \cite{chen2018ps,chen2019self, ju2020learning} applied the max-pooling method to aggregate features from a number of inputs. In addition, some works proposed to better handle specified problems. Taniai and Maehara \cite{taniai2018neural} proposed an unsupervised method to better handle the condition of lacking ground-truth surface normals, where the surface normals are estimated by minimizing the reconstruction loss. Ju \textit{et al.} \cite{ju2020pay} proposed an adaptive attention-weighted loss to improve the performance of complex-structured areas, where the detail-preserving gradient loss can produce clear reconstructions. More recently, Yao \textit{et al.} \cite{yao2020gps} attempt to introduce GNN for learning-based photometric stereo, named GPS-Net. Ju \textit{et al.} \cite{ju2021recovering} proposed a dual-regression method to recover both surface normal and rendered observations, which provides an additional supervision.

However, these deep learning methods still need to be improved when dealing with strong non-Lambertian surfaces and high-frequency structures. Unlike previous deep learning-based photometric stereo methods, we introduce the Lambertian priors to guide the learning and correct the errors caused by non-ideal solutions, which outperform state-of-the-art methods on challenging benchmark datasets. 

\section{Proposed Method}

In this section, we will present our proposed Lambertian priors photometric stereo network. Our goal is to measure surface normal of non-Lambertian surfaces with complex reflectance such as inter-reflections and cast shadows. Our strategy is to boost the measurement based on the physical Lambertian prior. Before introducing the network details, we will first present the learning objective.

\subsection{What Is Learned in Our Method}
\label{analysis}

Assuming that a surface point on the surface with a unit normal $\boldsymbol{n} \in \mathbb{R}^3$ is illuminated by $j^{th}$ light source with direction $\boldsymbol{l_j} \in \mathbb{R}^3$ in intensity observation $\boldsymbol{i_j} \in \mathbb{R}^3$. We artificially split the observation $\boldsymbol{i_j}$ into two parts: the diffuse $\boldsymbol{i_j^d}$ and the other $\boldsymbol{i_j^o}$, where $\boldsymbol{i_j}$ can be shown by $\boldsymbol{i_j^d}$ + $\boldsymbol{i_j^o}$. The $\boldsymbol{i_j^d}$ represents the ideal Lambertian reflectance, while the $\boldsymbol{i_j^o}$ stands for specularity, cast shadows, inter-reflections, and global illuminations. 

If we can exactly extract the diffuse $\boldsymbol{i_j^d}$ from observation $\boldsymbol{i_j}$, then the surface normal $\boldsymbol{n}$ can be expressed, according to ideal Lambertian photometric stereo \cite{Woodham1980Photometric}, as follows:
\begin{equation}
\label{eq3.1-1}
\boldsymbol{n}=\frac{\boldsymbol{L}^{-1} \boldsymbol{I^d}}{\lvert \boldsymbol{L}^{-1} \boldsymbol{I^d} \rvert}  \ , 
\end{equation}
where $\boldsymbol{I^d}$ = $[ \boldsymbol{i_1^d}, \boldsymbol{i_2^d}, \dots , \boldsymbol{i_j^d} ]'$ is the column vector of the diffuse part in $\{ 1, 2, \ldots , j \}$ and the matrix $\boldsymbol{L}$ = $[\boldsymbol{l_1}, \boldsymbol{l_2}, \dots , \boldsymbol{l_j}]'$ is composed of illumination direction in $\{ 1, 2, \ldots , j \}$.

However, few methods can exactly express the diffuse part from the observation, including outliers methods and illumination models. Rather than using sophisticated models or deep learning to further reduce fitting errors, we, hereafter, rethink the Lambertian model in the non-Lambertian condition. Here, we attend the prior surface normal $\boldsymbol{n'}$, which can also be calculated via the ideal physical model \cite{Woodham1980Photometric}:
\begin{equation}
\label{eq3.1-2}
\boldsymbol{n'}=\frac{\boldsymbol{L}^{-1} \boldsymbol{I}}{\lvert \boldsymbol{L}^{-1} \boldsymbol{I} \rvert}  \ ,
\end{equation}
where $\boldsymbol{I}$ = $[ \boldsymbol{i_1^d} + \boldsymbol{i_1^o}, \boldsymbol{i_2^d} +\boldsymbol{i_2^o}, \dots , \boldsymbol{i_j^d} + \boldsymbol{i_j^o} ]'$. It can be seen that the error of $\boldsymbol{n'}$ is caused by the non-Lambertian and other global effects $\boldsymbol{I^o}$ = $[ \boldsymbol{i_1^o}, \boldsymbol{i_2^o}, \dots , \boldsymbol{i_j^o} ]'$. Naturally, we wish to learn the nonlinear mapping by the pattern of observation $\boldsymbol{I}$ including $\boldsymbol{I^d}$ and $\boldsymbol{I^o}$, as follows:
\begin{equation}
\label{eq3.1-3}
\mathnormal{Mapping} \ ( \frac{\boldsymbol{L}^{-1} \boldsymbol{I}}{\lvert \boldsymbol{L}^{-1} \boldsymbol{I} \rvert} \to \frac{\boldsymbol{L}^{-1} \boldsymbol{I^d}}{\lvert \boldsymbol{L}^{-1} \boldsymbol{I^d} \rvert} \ , with \ \boldsymbol{I^d} + \boldsymbol{I^o} ). 
\end{equation}

\begin{figure*} [!htp]
  \includegraphics[width=1\textwidth]{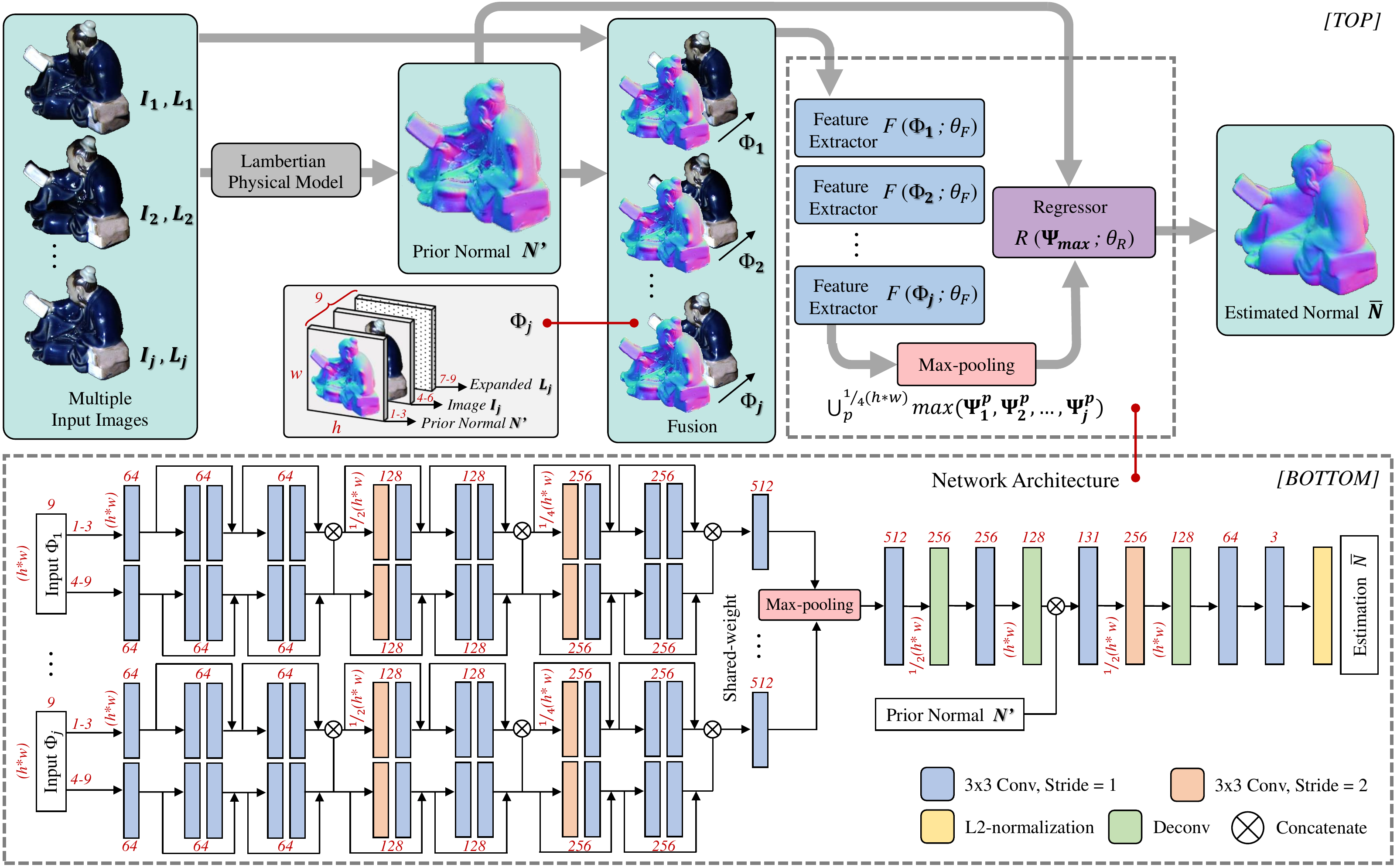}
  \caption{\textit{[TOP]} Overview. The prior normal $\boldsymbol{N}'$ is calculated by Lambertian physical model using input images $\{\boldsymbol{I_1},\boldsymbol{I_2},\ldots , \boldsymbol{I_j} \}$ with illumination directions $\{\boldsymbol{L_1},\boldsymbol{L_2},\ldots , \boldsymbol{L_j} \}$. Then, we fuse $\boldsymbol{N}'$ with $\{\boldsymbol{I_j}$, $\boldsymbol{L_j}\}$ in $\boldsymbol{\Phi_j}$, where the $\boldsymbol{L_j}$ is expanded to have the same spatial resolution as images. The detail of a $\boldsymbol{\Phi_j}$ is shown in the gray box, where the first three dimensions are normal and the last six dimensions are image and the expanded illumination direction. Our method takes the fusion $\{ \boldsymbol{\Phi_1}, \boldsymbol{\Phi_2}, \ldots , \boldsymbol{\Phi_j} \}$ including prior normals as inputs and estimates the surface normal $\boldsymbol{\bar{N}}$. \textit{[BOTTOM]} Network architecture. For each $\boldsymbol{\Phi}$, we split it into two parts, the prior normal and the image information. The channel dimension and spatial resolution of layers are highlighted in red.  }
  \label{fig2}
\end{figure*}

Unlike previous deep learning-based methods focusing on using different setups to infer the surface normal from solely input images, as $\mathnormal{Mapping} \ (  \boldsymbol{I^d} + \boldsymbol{I^o} \to \boldsymbol{n})$, our method focuses on the use of the input images $\boldsymbol{I^d} + \boldsymbol{I^o}$ as the patterns to correct the surface normal from Lambertian priors. For deep learning-based photometric stereo, we argue that the above mapping is a better choice for estimating the surface normal. Our analysis is that (1) mapping the errors in estimating surface normals $\{ \mathcal{Y} \to \mathcal{Y} \}$ instead of mapping the surface normals from RGB images $\{ \mathcal{X} \to \mathcal{Y} \}$ reduce the space of solving possible learning functions and improve the estimation results ($\mathcal{Y}$ represents the surface normal space, where $\mathcal{X}$ represent the RGB image space). (2) the prior normals $\boldsymbol{n'}$ are theoretically accurate in terms of diffuse reflection.In this condition, our method is equivalent to enlarge the proportion of non-Lambertian errors in total errors, where the network always tends to learn the parameters to make the loss drop. In training, the network therefore will be more inclined to optimize this part of errors. \emph{i.e.}, our method learns the differential features, amplifying the non-Lambertian errors.

\subsection{Network Architecture}

In this section, a novel Lambertian guided photometric stereo network is designed according to the inference reported in Section \ref{analysis}. The network enhances the measurement on non-Lambertian surfaces, crinkle regions, and varying reflectance edges. The overview and detailed architecture of our Lambertian priors based photometric stereo network is illustrated on the \textit{[TOP]} and  \textit{[BOTTOM]} of Fig. \ref{fig2}, respectively. 

We first fuse the Lambertian priors with input images. Given $j^{th}$ images with known illumination directions, we take the Lambertian assumption \cite{Woodham1980Photometric} to calculate the prior normals $\boldsymbol{N}'$ by Eq. \ref{eq3.1-2}. We then concatenate the prior normal with each observation and illumination as $\boldsymbol{\Phi_j}$. For the illumination of the corresponding image, we expand the 3-vector illumination and expand it to the same spatial resolution as the image, $\boldsymbol{L_j} \in \mathbb{R}^{3 \times h \times w}$, and concatenate the illumination $\boldsymbol{L_j}$ with the image. This operation makes the illumination directions fully fuse with the corresponding observation in a pixel-wise manner. Hence, each input $\boldsymbol{\Phi_j}$ has the dimensions of $\mathbb{R}^{9 \times h \times w}$, where the concatenation order is the prior normal, the image, and the illumination direction. In fact, the above mapping model (Eq. \ref{eq3.1-3}) is constructed in the per-pixel manner. However, the proposed network predicts the surface normal from the image patch. Inspired by previous works \cite{chen2018ps, wang2020non}, we argue that the embedded features from a neighboring image patch enhances the flexibility of the proposed network to various reflectance.  

We separately feed these $\{ \boldsymbol{\Phi_1}, \boldsymbol{\Phi_2}, \ldots , \boldsymbol{\Phi_j} \}$ to the network, as shown on the \textit{[BOTTOM]} of Fig. \ref{fig2}. The red numbers represent the dimensions of the feature channels. We apply the Leaky-ReLU as the activation function of each layer. The network includes three stages: feature extraction, max-pooling fusion, and regression.

The first stage of our network can be seen as the $j^{th}$ multi-branch shared-weight feature extraction, as:
\begin{equation}
\label{eq3.2-1}
\boldsymbol{\Psi_j} =F ( \boldsymbol{\Phi_j} ; \theta_F ),
\end{equation}
where $\boldsymbol{\Psi_j} \in \mathbb{R}^{512 \times \frac{1}{4} h \times \frac{1}{4} w} $ is the feature from the feature extractor $F$ with learnable parameters $\theta_F$. The feature extraction stage contains two residual networks \cite{he2016deep}, handling the surface normal and the image with its associated illumination direction, respectively. It can be seen that each residual network contains 6 residual blocks with two down-sampling convolutional layers. Residual blocks can effectively avoid gradient vanishing \cite{he2016deep} in a deep network. Also, we argue that the shortcut fuses previous blocks, which is a combination of features at different levels and scales. In addition, the shortcut structure is equivalent to adding all the information of the previous layer image in each block, which retains more original features. We also compare the results of different network architectures in ablation study (Section \ref{extractionstr}). The two down-sampling, from $h \times w$ to $ \frac{1}{4} h \times \frac{1}{4} w$ are executed by stride = 2 convolutional layer on the third and fifth residual blocks. By concatenating the features of the image residual network to those of the prior normal residual network in different scales, the network increase the receptive field. 

A convolutional layer is applicable for multi-feature fusion only when the number of inputs is fixed. Unfortunately, this is not practical for photometric stereo where the number of inputs often change. Therefore, in the second stage, we apply a max-pooling operation \cite{chen2018ps,ju2020pay} for multi-feature fusion, and getting fixed feature that can be backpropagated. Max-pooling extracts the most salient information from all the features. In fact, it is the regions with high intensities or specularities that provide strong clues for surface normal estiamtion. Also, the max-pooling operation can exclude the shadows from multi-illumination directions. In contrast, the average-pooling will smooth out useful features and may be impacted by non-activated features such as shadows. Here, we choose the superscript $p$ to denote the index of the feature tensor, as follows:
\begin{equation}
\label{eq3.2-2}
\boldsymbol{\Psi_{max}} = \bigcup_{p}^{\frac{1}{4}h \times \frac{1}{4} w} max ( \boldsymbol{\Psi_{1}^p}, \boldsymbol{\Psi_{2}^p}, \ldots , \boldsymbol{\Psi_{j}^p}). 
\end{equation}

Finally, the normal regression stage $R$ takes $\boldsymbol{\Psi_{max}}$ as input and regresses the estimation surface normal $\boldsymbol{\bar{N}}$ as:
\begin{equation}
\label{eq3.2-3}
\boldsymbol{\bar{N}} = R ( \boldsymbol{\Psi_{max}}  ; \theta_R ),
\end{equation}
where $\theta_R$ is the learnable parameter of the regressor $R$. The regressor consists of six convolutional layers, three deconvolutional layers, and an L2-normalization layer. The feature map is up-sampled (by the deconvolutional layer) three times and down-sampled (by the stride = 2 convolutional layer) once to fully utilize the embedded information, resulting in up-sampling of twice. We argue that this design can expand the receptive field and keep spatial information with a small GPU memory. We concatenate the prior normal $\boldsymbol{N}'$ to the feature map after the second deconvolutional layer. We reckon that the fusion in the regression stage will enhance the high-frequency details of the estimated surface normal. The detailed discussion can be found in the ablation experiments (Section \ref{ablation}).

The learning of our network is supervised by the error between the measured and the ground-truth surface normals. We optimize the networks parameters $\theta_{F}$, $\theta_{R}$ by minimizing the cosine similarity loss function as:
\begin{equation}
\label{eq3.2-4}
\mathcal{L}_{\mathrm{normal}} = \frac{1}{h w} \sum_{p}^{h w}\left(1-\boldsymbol{\bar{N}}^{p} \cdot \boldsymbol{N}^{p}\right),
\end{equation}
where $\boldsymbol{\bar{N}}^{p}$ and $\boldsymbol{N}^{p}$ denote the measured  normal and the ground-truth, respectively, at pixel $p$. If the estimated normal $\boldsymbol{\bar{N}}^{p}$ at pixel $p$ has a similar orientation as the ground-truth $\boldsymbol{N}^{p}$, then the $\boldsymbol{\bar{N}}^{p} \cdot \boldsymbol{N}^{p}$ will be close to 1 and the cosine similarity loss will approach 0.

Our network was implemented using PyTorch and the Adam optimizer is used with the default settings ($\beta_1$= 0.9 and $\beta_2$=0.999) on a RTX 2080 GPU. The initial learning rate is set to 0.002, and divided by 2 every 5 epochs.  We train the model using a batch size of 24 for 40 epochs. The number of observations for training and prior normals is 32. In addition, we set the spatial resolution $h, w$ = 32 in training.

\section{Experiments}

In this section, we present datasets, experimental results, and analysis. To verify the quantitative performance of our method, we use some widely used metrics to measure accuracy. We adopt the mean angular error (MAE) in degree to evaluate the performance of estimated surface normal, as follows:
\begin{equation}
\label{eq4-1}
MAE = \frac{1}{HW} \sum_{p}^{H \times W}(arccos(\boldsymbol{\bar{N}}^p  \cdot \boldsymbol{N}^p)) \ ,
\end{equation}
where $H \times W$ represents the spatial resolution of the tested surface normal. We also apply $< err_{{15}^\circ}$ and $< err_{{30}^\circ}$ to measuring the percentage (\%) that pixels with the angular error less than $15^\circ$ and $30^\circ$, respectively.

\subsection{Dataset}
\subsubsection{Training and validation datasets}

As a supervised learning method, we adopt two publicly available synthetic datasets from \cite{chen2018ps}, called blobby shape and sculpture shape datasets \cite{johnson2011shape}, which are rendered with the MERL dataset \cite{matusik2003data} by the physically-based raytracer Mitsuba \cite{jakob2010mitsuba}. Blobby and Sculpture datasets provide surfaces with complex structures and rich surface orientations, and the MERL dataset contains 100 different BRDFs of real-world materials. The combinations provide comprehensive data distribution. The training set contains 85212 samples. For each sample, 64 observation images are rendered by random illumination directions in an upper-hemisphere, with a 128 $\times$ 128 spatial resolution. We randomly crop 32 $\times$ 32 (default training spatial resolution) images patches in each sample for data augmentation.

\subsubsection{Testing datasets}

To evaluate our method, we apply several commonly used datasets, including both synthetic and real datasets. For the synthetic dataset, we first employ the CyclesPSTest dataset \cite{ikehata2018cnn}. CyclesPSTest is a synthetic dataset of three objects, ``Sphere'', ``Turtle'', and ``Paperbowl''. ``Turtle`` and ``Paperbowlare'' are objects with the non-convex surface where specularity and cast shadow extensive appear. We also employ the synthetic object ``Dragon''. The object ``Dragon'' was rendered with 100 different BRDFs from the MERL data set \cite{matusik2003data} under 100 random illumination directions in an upper-hemisphere, for testing the results of our method on different surface reflectances.

For the real dataset, we employ the public DiLiGenT benchmark dataset \cite{shi2019benchmark}, which contains 10 objects of various shapes with complex materials. Each object provides images with a resolution of 612 $\times$ 512 from 96 different known illumination directions. The DiLiGenT benchmark dataset is challenging for its strong non-Lambertian surfaces and non-convex structures. Besides, we also employ the Light Stage Data Gallery \cite{einarsson2006relighting}, which contains six objects without ground-truth. Each object has 253 images under different illumination directions. Therefore we qualitatively evaluate our method on the Light Stage Data Gallery.

\subsection{Ablation Experiments and Network Analysis}
\label{ablation}
\begin{figure*} [!htp]
  \centering\includegraphics[width=1\textwidth]{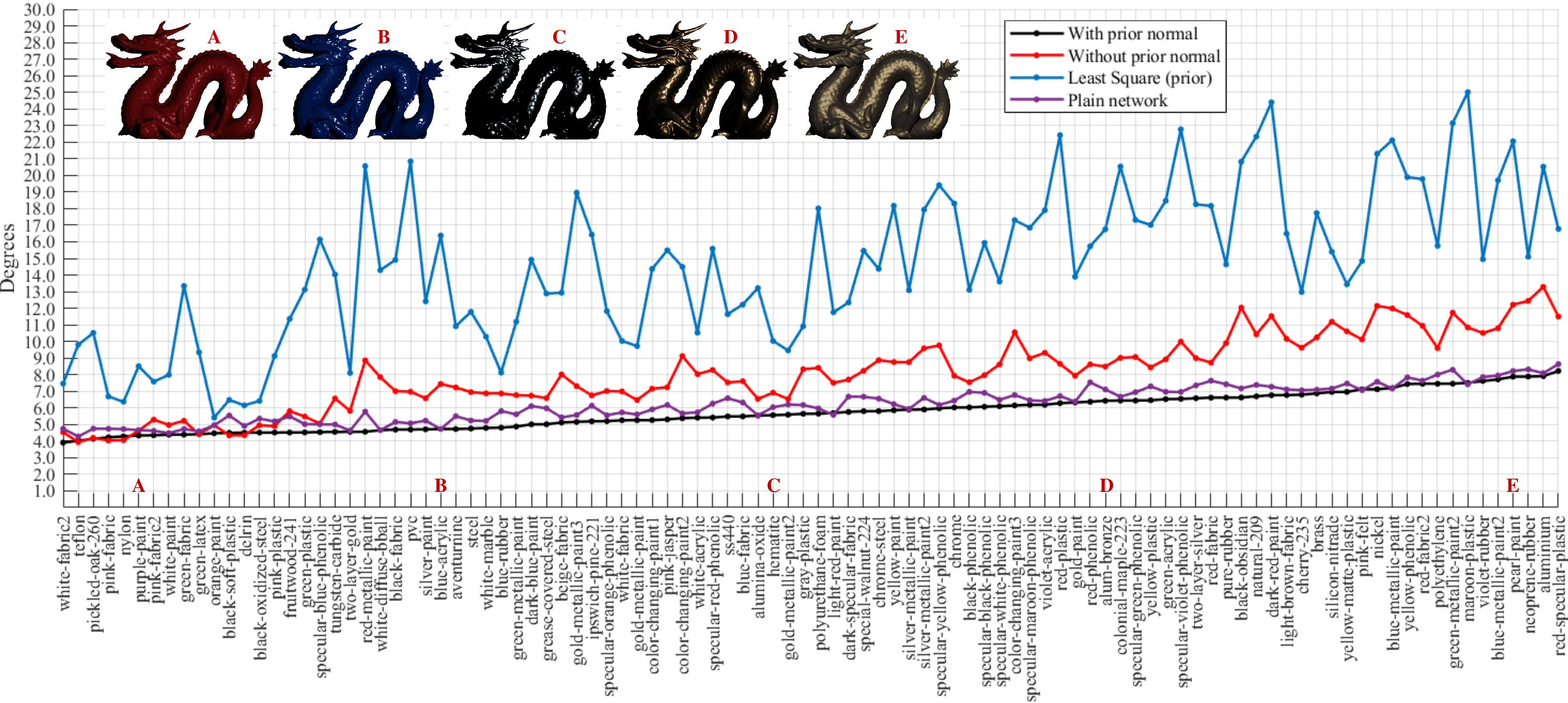}
  \caption{The MAE in degrees of the estimated surface normals on the samples of ``Dragon'' with 100 materials from MERL BRDFs \cite{matusik2003data}. We report the performance of our default settings, without prior normal, plain network (with prior normal), and least square method (the prior normal) \cite{Woodham1980Photometric}. }
  \label{fig3}
\end{figure*}

We took quantitative ablation experiments on the validation set of 852 samples. We first evaluated the effectiveness of our Lambertian priors based photometric stereo network (Experiments with IDs 0 \& 1). We compared our default network with only using input images, where the residual network for the Lambertian priors branch (channel 1-3 of each input $\boldsymbol{\Phi_j}$) and concatenation are removed. We then investigated the influence of different prior inputs on our network (Experiments with IDs 0 \& 2). For comparisons, we selected rank minimization \cite{wu2010robust} (a robust photometric stereo method) as the prior input. Furthermore, we evaluated the effectiveness of the selected residual architecture (experiments with IDs 0 \& 3). We compared the residual blocks’ settings \cite{he2016deep} in the feature extraction stage with the plain settings (without shortcut connections). Finally, we evaluated the effectiveness of concatenating the prior normal $\boldsymbol{N}'$ in the regression stage (experiments with IDs 0, 4, \& 5). We compared the concatenation with the prior after the second deconvloutional layer (default) with the concatenation with the prior after the third deconvloutional layer, and without the concatenation with the prior. For all the experiments in the ablation study, we train the networks with 32 input images, and reported the average results of the validation set of 852 samples. The results were summarized in Table \ref{tab1}.  To further evaluate the performance and the generalization of our method, we also report the ablation study on the synthetic object ``Dragon'' with 100 different materials, as shown in Fig. \ref{fig3}, corresponding to IDs 0, 1, and 3 in Table \ref{tab1}.

\begin{table}[!htp]
\caption{Results of system analysis on the validation set with 32 input images, where the numbers represent the average values of all the samples of the validation set. The lower MAE, the better.  For $< err_{{15}^\circ}$ and  $< err_{{30}^\circ}$, the higher, the better.}
\label{tab1}
\begin{center}
\begin{tabular}{llccc}
  \hline 
  ID &Variants  &  MAE  &$< err_{{15}^\circ}$  &$< err_{{30}^\circ}$ \\ 
  \hline 
  
   0& With prior normal (Ours) & 12.30& \textbf{84.39\%}&94.86\% \\
   \hline 
  1 & Without prior normal & 12.98& 82.05\%& 94.71\%\\
  
  2& With rank minimization \cite{wu2010robust}&\textbf{12.27}&84.26\%&\textbf{94.91\%}\\
  
  3& Plain network & 12.45&83.94\%&94.83\%\\
  4& Concatenate 3rd deconv & 12.32&84.15\%&94.65\%\\
  5& Without concatenate & 12.54&82.85\%&94.61\% \\
  \hline 
\end{tabular}
\end{center}
\vspace{-3mm}
\end{table}%

\subsubsection{Effectiveness of Lambertian priors based photometric stereo network} 

Experiments with IDs 0 and 1 demonstrated the performance of adding Lambertian priors to the photometric stereo networks. It can be seen that our method of using prior normal consistently performs better than the traditional mapping strategy over all the metrics, for example, $12.30^\circ$ for MAE, 84.39\% and 94.86\% of the pixels have the angular errors of less than $15^\circ$ and $30^\circ$. This is due to the fact that our method learns the mapping in the same normal space while the previous methods learn the mapping over different spaces: from RGB images to normal. Therefore our method can converge faster and achieve accurate estimations. Also, as shown in Fig. \ref{fig3}, It can be seen that on most materials, our method significantly outperformed the network without a normal prior and the baseline Lambertian method \cite{Woodham1980Photometric}. Note that the proposed method performed particularly well on the surfaces that have larger errors in the baseline method, which suggested our method can improve the prediction over strong non-Lambertian reflectance. 

\begin{figure} [!htp]
  \centering\includegraphics[width=0.48\textwidth]{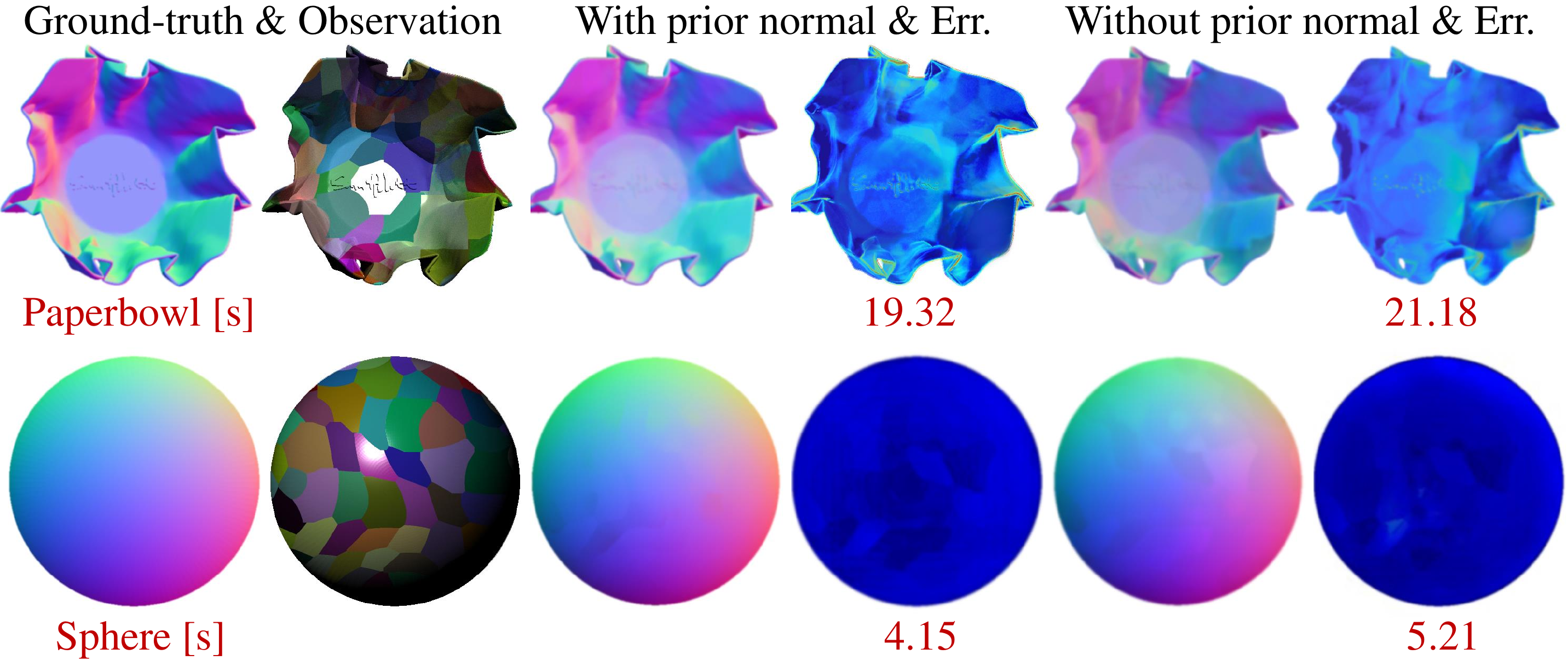}
  \caption{Visual comparisons between networks with Lambertian priors and without prior normal. We show the results of ``Paperbowl'' and ``Sphere'' in synthetic CyclesPSTest dataset \cite{ikehata2018cnn} with only 17 input images, [s] stands for Specular.}  
  \label{fig4}
  \vspace{-5mm}
\end{figure}

We further evaluated the robustness of our method with spatially varying BRDFs on the surface. Fig. \ref{fig4} futher quantitatively shows two objects from ``Paperbowl'' and ``Sphere'' in synthetic CyclesPSTest dataset \cite{ikehata2018cnn} with only 17 input images. Similar to the results on the synthetic object ``Dragon'', our method outperformed the counterpart without prior normal. It can be seen that reflectance is rapidly changed on these two object, denote that our method can lead to smoother surface normals compared with the method of using only image input which suffer from wide varieties of real-world materials.

\begin{figure} [!htp]
  \includegraphics[width=0.48\textwidth]{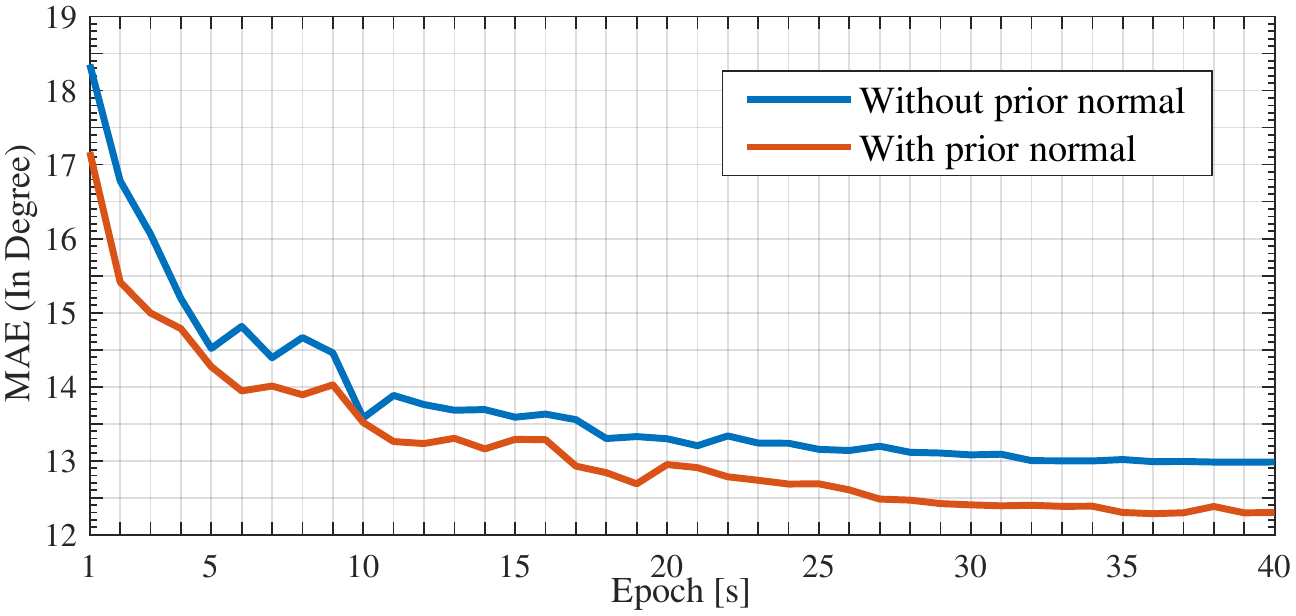}
  \caption{Convergence comparison on the validation set. The blue curve represents the network without the prior normal while the orange curve is the network using Lambertian priors. For the network without the prior normal, we keep the same architecture but remove the second residual network which handles the prior normal, and all the concatenate operations. Both two networks are trained with the same parameters and 32 images as the input. }
  \label{fig5}
  \vspace*{-5mm}
\end{figure}

In addition, we reported the convergence error in Fig. \ref{fig5}. As shown in Fig. \ref{fig5}, our method of using Lambertian priors achieved lower convergence error in the training processing. It shows that our network is more effective for feature extraction and regression than the previous mapping method learning surface normal from only input images.  

Moreover, the convergence of our method is faster, which means that our method can further benefit from fewer training samples condition. To prove this, we tabulated the performance with fewer training datasets in Table \ref{datanumber}. The training with fewer dataset causes larger errors on methods whether using Lambertian priors. However, our method with priors performs slight decrease ($0.34^\circ$ ) when using Blobby dataset with 25600 samples, while the method without priors reports a larger drop ($0.81^\circ$). In fact, if our method was trained by very few samples, then the error would be close to least square priors \cite{Woodham1980Photometric}. However, the errors of the method without priors might be unacceptable.

\begin{table}[!htp]
\caption{Normal estimation results  on the validation set with 32 input images. Blobby dataset contains 25660 training samples while Sculpture gives 58700 samples.}
\label{datanumber}

\begin{center}
\begin{tabular}{lccc}
  \hline 
  Training datasets  &  With prior normal  & without prior normal \\ 
  \hline 
  Only Blobby &\textbf{12.64}&13.79 \\
  Only Sculpture& \textbf{12.39}&13.23\\
  Blobby + Sculpture&\textbf{12.30} &12.98\\
  \hline 
\end{tabular}
\end{center}
\vspace{-3mm}
\end{table}%

We also observed that the MAE of the validation set is larger than that of the DiLiGenT dataset (see in Section \ref{diligent}). The reason for this may be because the illumination directions of the validation set are randomly generated in an upper-hemisphere, while the illumination directions of the DiLiGenT dataset are more clustered, which benefits the learning with fewer cast shadows.

\subsubsection{Effectiveness of different priors} 

Experiments with IDs 0 and 2 show the influence of different prior inputs on our network. In ID 2, we use the results of the rank minimization method \cite{wu2010robust} as the prior normal. Referring to Table \ref{tab1}, we discover that the prior normal from the rank minimization is slightly better than that from the Lambertian model on MAE and $< err_{{30}^\circ}$. This is due to the fact that the rank minimization method \cite{wu2010robust} has better performance than the Lambertian model. However, we still choose the Lambertian surface normal as the priors because of the multi-facet. First, the rank minimization method needs much time for detecting and removing the outliers. Therefore, the time consumption of the network using rank minimization priors is demanding on both the training (65 hours for training, while Lambertian priors only need 19 hours) and the test stages. Second, the rank minimization priors are limited in varying reflectance surfaces, because the outlier removal methods such as rank minimization are generally effective for limited categories of reflectance.

\subsubsection{Effectiveness of residual blocks in the feature extraction stage}
\label{extractionstr}
Experiments with IDs 0 and 3 show the performance of residual architectures and plain counterparts in the feature extraction stage. Referring to Table \ref{tab1}, experiments show that applying residual blocks in the feature regression stage has a lower mean angular error. We also found that the $< err_{{15}^\circ}$ had a relatively large drop when the plain network was used. Also, as shown in Fig. \ref{fig3}, It can be seen that our method has a slightly small error among most materials and average MAE. The results suggested the residual architecture increased the accuracy of surface normal estimation. The reason might be because the residual blocks can effectively avoid gradient vanishing \cite{he2016deep} in a deep network. Also, we argue that the shortcut fuses previous blocks, which is a combination of features at different levels and scales. In addition, the shortcut structure is equivalent to adding all the information of the previous layer image in each block, which retains more original features.

We further compare the different architectures of the residual module. As tabulated in Table \ref{residualnumber}, our default settings (6 residual blocks) achieve better performance compared with using fewer residual blocks. We find that our default settings are slightly worse than 7 residual blocks on MAE. However, the additional residual blocks increase the parameters and training time. For simplifying the complexity of the network, we just remain the 6 blocks eventually. Also, we compare our method with a more simplified single residual branch (directly handle all channels of input, rather than handle the 1-3 channels and 4-9 channels by two residual branches, as shown in Fig. \ref{fig2} \textit{[BOTTOM]}). However, the performance of using the single residual branch is worse.

\begin{table}[!htp]
\caption{Results of the different residual blocks architectures testing on the validation set.}
\label{residualnumber}
\begin{center}
\begin{tabular}{lcccc}
  \hline 
 Architectures &  MAE  & $< err_{{15}^\circ}$&$< err_{{30}^\circ}$ \\ 
  \hline 
  ours (6 residual blocks) &12.30&\textbf{84.39\%} &\textbf{94.86\%} \\
  3 residual blocks& 12.37&84.35\% &94.83\% \\
  5 residual blocks&12.32 &84.38\% &94.85\% \\
  7 residual blocks&\textbf{12.29} & \textbf{84.39\%} & 94.85\%\\
  Single branch&12.52 &84.21\% &84.77\% \\
  \hline 
\end{tabular}
\end{center}
\vspace*{-2mm}
\end{table}%

\subsubsection{Effectiveness of concatenating prior normal} 

Experiments with IDs 0, 4, and 5 shown in Table \ref{tab1} reported the effectiveness of concatenating prior normals in the regression stage. For ID 4, we concatenated the prior normal to the third deconvolutional layer in the regression stage. For ID 5, we removed the concatenation operation in the regression stage (without any prior normal). Table \ref{tab1} shows that the default settings (concatenating the prior normal to the second deconvolutional layer) helped us to boost the performance. In particular, non-concatenating in the regression stage has a negative impact on the predicted results. This suggests that adding prior normals in deep feature layers will enrich details and increase accuracy. We also noted that the performance was slightly worse, when moving the concatenation back to the 3rd deconvolutional layer. This may be due to the fact that the subsequent up-sampling and down-sampling operations well support the feature fusion in different scales.

\subsection{Evaluation on the DiLiGenT benchmark dataset}
\label{diligent}

\begin{table*}[!htb]
\begin{center}
\caption{Comparison of different methods on the DiLiGenT benchmark dataset. All methods are evaluated with 96 images. Here, we measure MAE in degrees. Our method was trained with 10, 32, 64 images, respectively. Black bold texts represent the best performance, and underlined texts represent the second best. } 
\label{tab2}
\begin{tabular}{l c c c c c c c c c c c }
\hline
Method & Ball&Bear& Buddha &Cat& Cow&Goblet&Harvest&Pot1&Pot2&Reading&Avg. \\
\hline
Baseline (Least squares) \cite{Woodham1980Photometric}&4.10&8.39&14.92&8.41&25.60&18.50&30.62&8.89&14.65&19.80&15.39\\
Monotonic BRDF \cite{shi2012elevation}&13.58&19.44&18.37&12.34&7.62&17.80&19.30&10.37&9.84&17.17&14.58 \\
Matrix rank = 3 \cite{ikehata2012robust}&2.54&7.32&11.11&7.21&25.70&16.25&29.26&7.74&14.09&16.17&13.74\\
Rank minimization \cite{wu2010robust}&2.06&6.50&10.91&6.73&25.89&15.70&30.01&7.18&13.12&15.39&13.35\\
Consensus constraint \cite{higo2010consensus}&3.55&11.48&13.05&8.40&14.95&14.89&21.79&10.85&16.37&16.82&13.22\\
2D discrete table \cite{alldrin2008photometric}&2.71&5.96&12.54&6.53&21.48&13.93&30.50&7.23&11.03&14.17&12.61 \\
Multi-Ward models \cite{goldman2009shape}&3.21&6.62&14.85&8.22&9.55&14.22&27.84&8.53&7.90&19.07&12.00\\
Bivariate BRDF \cite{ikehata2014photometric}&3.34&7.11&10.47&6.74&13.05&9.71&25.95&6.64&8.77&14.19&10.60 \\
Bi-polynomial \cite{shi2014bi}& \underline{1.74}&6.12 &10.60 &6.12 &13.93 &10.09 &25.44 &6.51 &8.78 &13.63 &10.30  \\
SDPS-Net \cite{chen2019self}&2.77&6.89&8.97&8.06&8.48&11.91&17.43&8.14&7.50&14.90&9.51\\
DPSN \cite{santo2017deep}&2.02&6.31&12.68&6.54&8.01&11.28&16.86&7.05&7.86&15.51&9.41 \\
IRPS \cite{taniai2018neural} &\textbf{1.47} &5.79& 10.36 &5.44&6.32 &11.47 &22.59 &6.09 &7.76 & \underline{11.03} &8.83 \\
PS-FCN \cite{chen2018ps} &2.82 &7.55 &7.91 &6.16 &7.33 &8.60 &15.85 &7.13 &7.25 &13.33 &8.39 \\
CNN-PS \cite{ikehata2018cnn}&2.12&12.30 &8.07&\textbf{4.38}&7.92&\textbf{7.42}&\textbf{13.83}&\textbf{5.37}&\textbf{6.38}&12.12&7.99 \\
Attention-PSN \cite{ju2020pay}&2.93&\textbf{4.86}&\underline{7.75}&6.14&6.86&8.42&15.44&6.92&\underline{6.97}&12.90&7.92 \\
DR-PSN \cite{ju2021recovering} &2.27& 5.46& 7.84& \underline{5.42}& 7.01&8.49& 15.40& 7.08&7.21& 12.74& 7.90 \\
GPS-Net \cite{yao2020gps}&2.92&\underline{5.07}&7.77&\underline{5.42}&\textbf{6.14}&9.00&15.14&\underline{6.04}&7.01&13.58&\underline{7.81}\\
\hline
Ours (Trained with 10 images)&2.57&5.89&8.94&7.10&7.54&8.82&15.48&7.68&7.71&11.53&8.33\\
Ours (Trained with 64 images)&2.49&5.64&\textbf{7.70}&6.45&\underline{6.23}&\underline{8.36}&\underline{14.67}&7.12&7.22&\textbf{10.89}&\textbf{7.68}\\
\hline
Ours (Trained with 32 images, default)&2.51&5.77&7.88&6.56&\underline{6.29}&\underline{8.40}&\underline{14.95}&7.21&7.40&\textbf{11.01} &\textbf{7.80}\\
\hline
\end{tabular}
\end{center}
\vspace*{-2mm}
\end{table*}

\begin{table*}[!htb]
\begin{center}
\caption{Comparison of different methods on the DiLiGenT benchmark dataset. We note that all methods are evaluated with 10 images for MAE in degrees. Our method was trained with 10, 32, 64 images, respectively. Black bold texts represent the best performance, and underlined texts represent the second best.} 
\label{tab3}
\begin{tabular}{l c c c c c c c c c c c }
\hline
Method & Ball&Bear& Buddha &Cat& Cow&Goblet&Harvest&Pot1&Pot2&Reading&Avg. \\
\hline
Baseline (Least squares) \cite{Woodham1980Photometric}& 5.09& 11.59 &16.25 &9.66 &27.90&19.97 &33.41 &11.32 &18.03 &19.86 &17.31\\
Bi-polynomial \cite{shi2014bi}& 5.24&9.39&15.79&9.34&26.08&19.71&30.85&9.76&15.57&20.08&16.18\\
Matrix rank = 3 \cite{ikehata2012robust}&\textbf{3.33}&7.62&13.36&8.13&25.01&18.01&29.37&\underline{8.73}&14.60&16.63&14.48\\
CNN-PS \cite{ikehata2018cnn}&9.11& 14.08 &14.58 &11.71 &14.04 &15.48 &19.56 &13.23 &14.65 & 16.99 &14.34  \\
PS-FCN \cite{chen2018ps} &4.02 &\underline{7.18} &9.79 &8.80 &10.51 &11.58 &18.70 &10.14 &9.85 &15.03 &10.51  \\
SPLINE-Net \cite{zheng2019spline}& 4.96&\textbf{5.99}&10.07&\underline{7.52}&\underline{8.80}&10.43&19.05&8.77&11.79&16.13&10.35\\
LMPS \cite{li2019learning} & 3.97 &8.73 &11.36 &\textbf{6.69} &10.19 & 10.46 &17.33 &\textbf{7.30} &9.74 & \underline{14.37} &10.02 \\
DR-PSN  &3.83 &7.52&\textbf{9.55} &7.92&9.83& \underline{10.38} &\underline{17.12} &9.36 &\textbf{9.16} &14.75 &\underline{9.94} \\
\hline
Ours (Trained with 10 images) &\underline{3.62}&7.36&\underline{9.61}&7.66&\textbf{8.42}&\textbf{10.17}&\textbf{16.70}&9.24&\underline{9.38}&\textbf{14.15}&\textbf{9.63}\\
Ours (Trained with 64 images)& 3.94& 7.60&9.83& 7.94& \textbf{8.63}& \underline{10.38}& \textbf{17.07}& 9.45& \underline{9.73}& 14.42& \underline{9.94}\\
\hline
Ours (Trained with 32 images, default)&3.86&7.49&\underline{9.69}&7.82&\textbf{8.55}&\textbf{10.31}&\textbf{16.94}&9.28&\underline{9.54}&\textbf{14.30}& \textbf{9.78}\\
\hline
\end{tabular}
\end{center}
\vspace*{-2mm}
\end{table*}

\begin{figure*} [!htb]
  \includegraphics[width=1\textwidth]{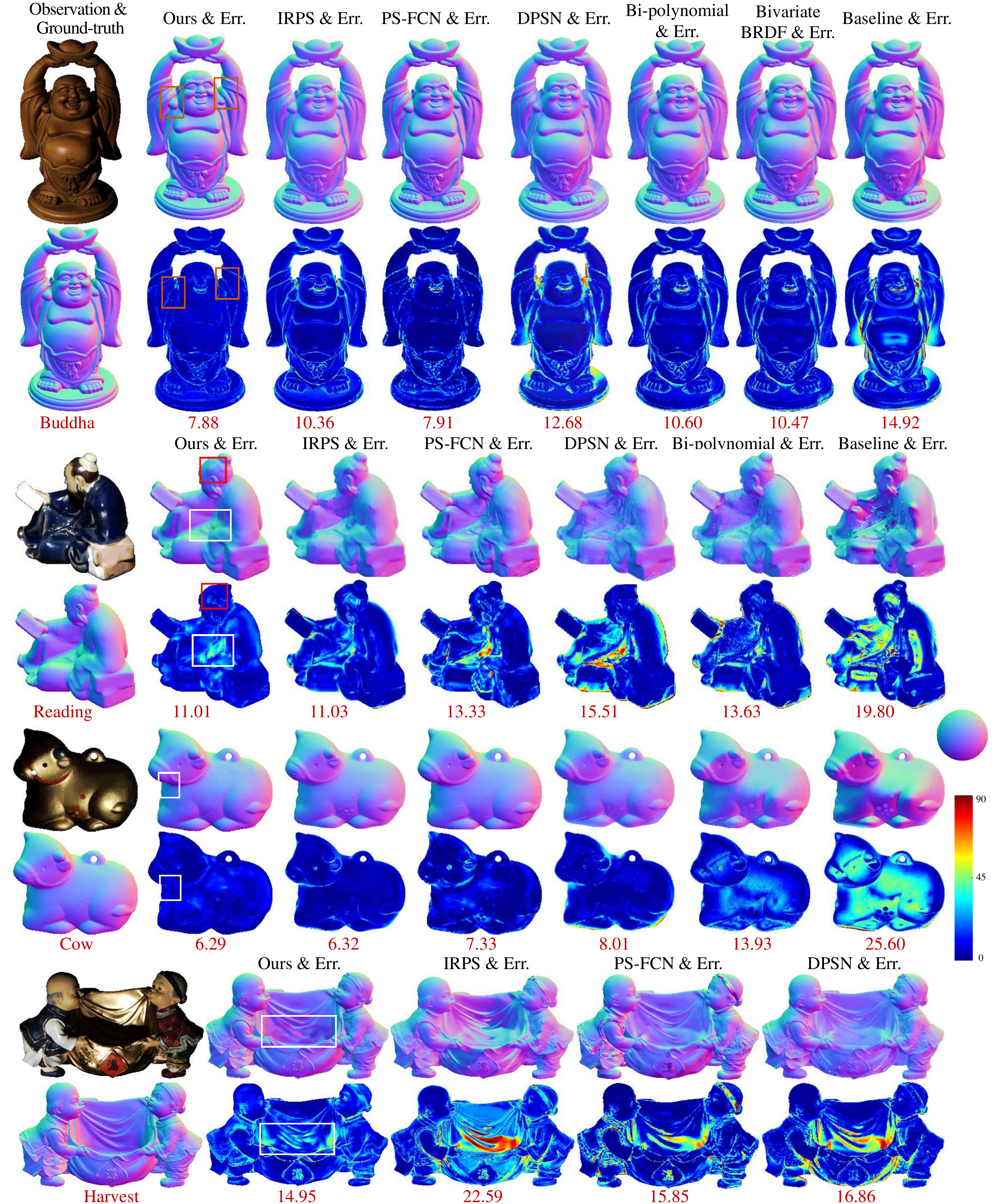}
  \caption{Quantitative results for strong non-Lambertian real-world scenes from the  DiLiGenT benchmark dataset. Err. is short for error map. The numbers under the error maps represent MAE in degrees. The contrast of observations are adjusted for easy view. The red boxes are regions with specularities, the white boxes are the regions with cast shadows, and the orange boxes represent the region with non-convex surfaces (crinkles). Our method produces more accurate estimations in those regions compared with the other methods. }
  \label{fig6}
\end{figure*}

We reported the results on the DiLiGenT benchmark dataset \cite{shi2019benchmark} with 96 input images in Table \ref{tab2}. In Table \ref{tab2}, we compared our Lambertian model guided photometric stereo network with both traditional algorithms and deep learning methods. For traditional algorithms, we evaluate the Lambertian baseline (our prior normal) \cite{Woodham1980Photometric}, robust methods \cite{ wu2010robust, ikehata2012robust} of outlier rejection-based technologies (robust method), and analytic and empirical models \cite{goldman2009shape,alldrin2008photometric, ikehata2014photometric, higo2010consensus,shi2014bi,shi2012elevation}. For deep learning methods, we also compared with several state-of-the-art networks, such as DPSN \cite{santo2017deep}, IRPS \cite{taniai2018neural}, PS-FCN \cite{chen2018ps}, CNN-PS \cite{ikehata2018cnn}, Attention-PSN \cite{ju2020pay}, DR-PSN \cite{ju2021recovering}, and GPS-Net \cite{yao2020gps}. Besides, we also evaluated our method against several deep learning methods with sparse inputs (with 10 input images) \cite{ikehata2018cnn,li2019learning,zheng2019spline} and flexible inputs methods \cite{chen2018ps,Woodham1980Photometric,ikehata2012robust,shi2014bi,ju2021recovering} shown in Table \ref{tab3}. Note that LMPS method \cite{li2019learning} takes 10 optimal images as inputs, while other sparse methods takes 10 random images as input. In particular, we reported our method of training with 10, 32, and 64 input images, respectively.  

\begin{figure*} [!htp]
  \includegraphics[width=1\textwidth]{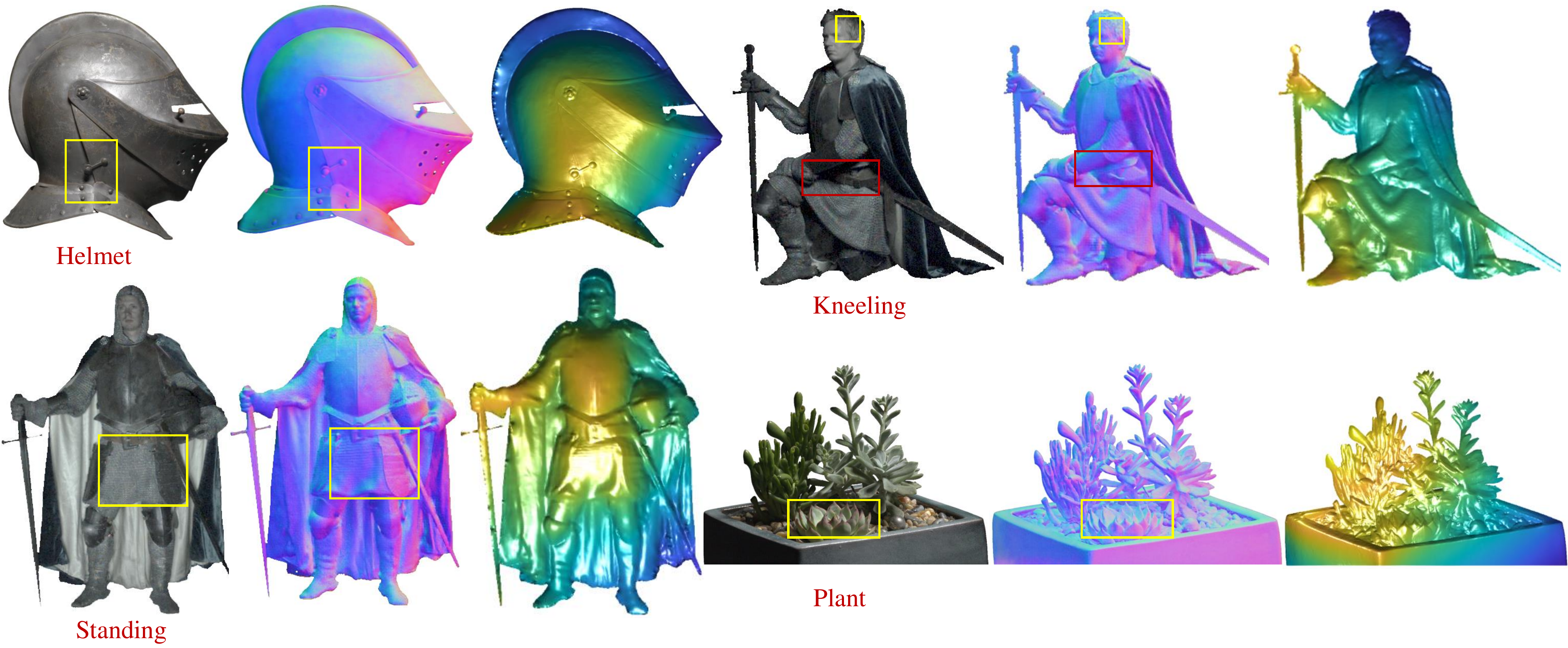}
  \caption{ Qualitative results of our method on objects ``Helmet'',  ``Kneeling'', ``Standing'', and ``Plant''. The contrast of observations is adjusted for easy viewing. For each object, the surface normal estimation and 3D reconstruction using \cite{simchony1990direct} are shown after the observation. The yellow boxes are the regions with complex structures, while the red boxes represent the regions with cast shadows. }
  \label{fig7}
  \vspace*{-2mm}
\end{figure*}

\begin{figure*} [!htb]
    \centering\includegraphics[width=0.75\textwidth]{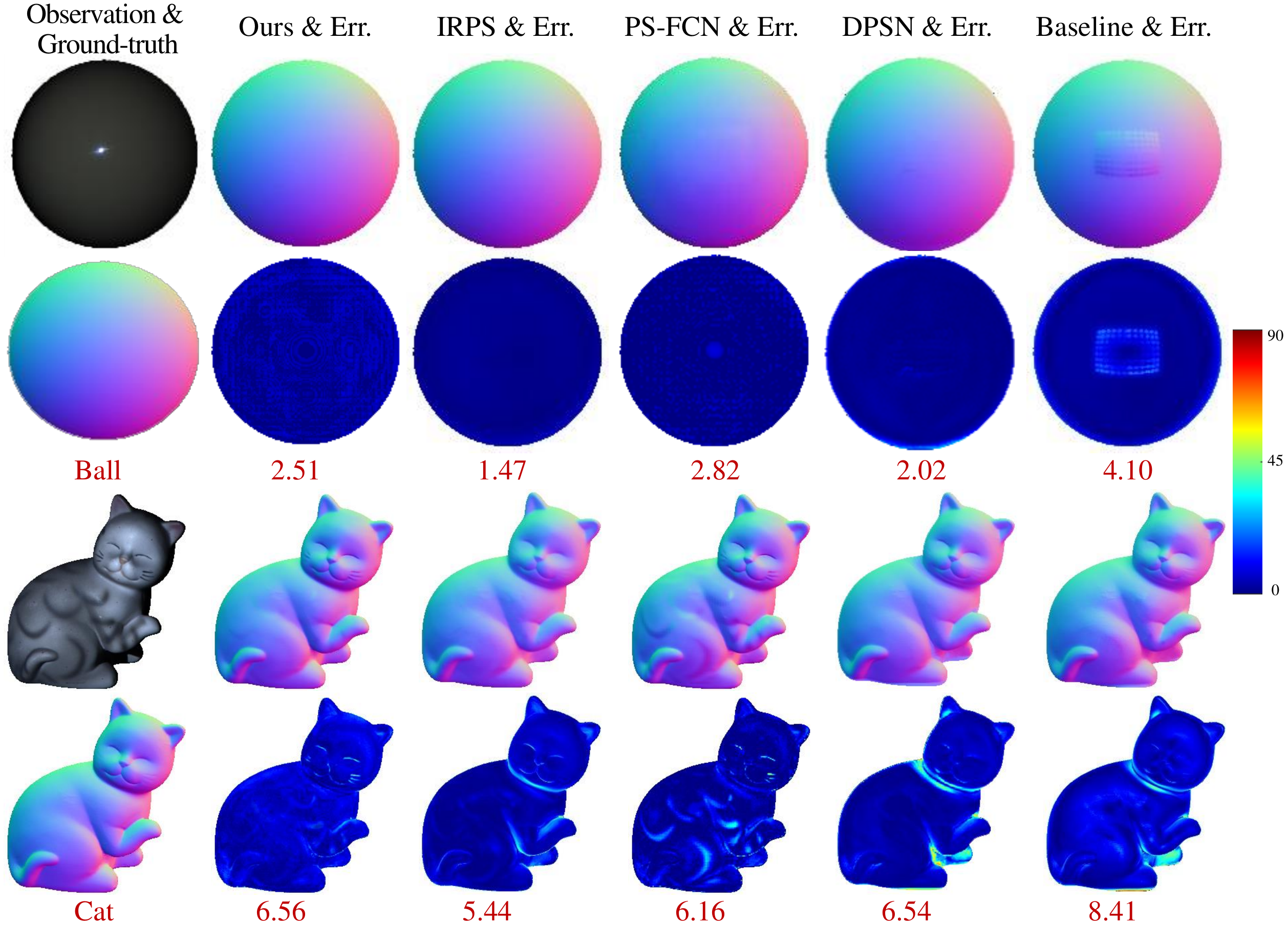}
  \caption{Quantitative results on``Ball'' and ``Cat'' from the  DiLiGenT benchmark dataset. Err. is the abbreviation for the error map. The numbers under the error maps represent MAE in degrees. The contrast of the observations is adjusted for easy view. }
  \label{fig8}
  \vspace*{-2mm}
\end{figure*}

As shown in Tables \ref{tab2} and \ref{tab3}, our method (default) outperformed the other state-of-the-art methods on both 96 and 10 input images, with higher average MAE.  In particular, it can be seen that our prior guided photometric stereo network substantially improves the cases with strong non-Lambertian and non-convex surfaces, such as ``Buddha'', ``Cow'', ``Harvest'', and ``Reading''. Therefore, we can reveal that providing prior normal has the effectiveness of handling specularity and cast shadows. We compared and showed the results of these objects in Fig. \ref{fig6}. It can be observed that our method recovered accurate surface normals on the regions with specularity, cast shadows, and crinkles, such as the collar of the ``Buddha'', the wrinkled clothes of the ``Reading'', and the pocket of the ``Harvest''. Note that the original CNN-PS \cite{ikehata2018cnn} discarded the first 20 images of “Bear” (tested with 76 images remaining), achieving a much better MAE of $4.25^{\circ}$ for “Bear”. The reason for discarding first 20 images of ``Bear'' is that the intensity values around the stomach region are wrong. However, all the images of ``Bear'' were used in evaluating all the other methods, except CNN-PS. For a fair comparison, we show the results based on the same test images (all 96 images).

In addition, we furthermore explored the influence of different input images in training. As shown in Tables \ref{tab2} and \ref{tab3}, we reported the results of our method trained with 10 and 64 images respectively (64 is the maximum number of the images in the training dataset). It can be seen that the method trained with 64 images achieved even better performance than the one trained with 32 images when using 96 images. On the contrast, the method trained with 10 images outperformed the method when using 10 images. In other words, similar input images between the training and the evaluation will benefit the estimation of the surface normal. The reason is that the prior normal is related to the input images, where the differences of prior normals caused by varying numbers of input images will affect the patterns the method learned to some extent. In order to obtain the best performance, we, therefore, recommend that a similar number of input images be used during the training and testing. Nonetheless, our default setting (trained with 32 images) has achieved the state-of-the-art performances on the measurement with 96 and 10 images (for a fair comparison).

\subsection{Evaluation on the Light Stage Data Gallery}
We further evaluated our method on a more complex dataset with general non-Lambertian materials. Fig. \ref{fig7} shows the results of our method using the Light Stage Data Gallery \cite{einarsson2006relighting}. We show the qualitative  outcomes for four complex objects ``Helmet'', ``Kneeling'', ``Standing'', and ``Plant'' in the dataset, due to the absence of ground-truth surface normals. Similarly, our method was trained with 32 images while being evaluated with random 96 observations in all 253 images.

As shown in Fig. \ref{fig7}, the estimated normal keeps the details without blur, such as the screws of the ``Helmet'', the hair of the ``Kneeling'', the lumpy-looking clothes of the ``Standing'', and the succulent plants of the ``Plant''. Note that the reflectance of plants was not trained in the training process. However, the result of object ``Plant'' is quite visually accurate, which shows the robustness of our method. We can also see the accurate reconstruction of the cast shadow areas (red boxes in Fig .\ref{fig7}). The reconstructed surface normal and the integral mesh convincingly reflect the shapes of the objects, demonstrating the accuracy of our physical prior photometric stereo network. We also observed that the estimated surface normal of the same objects, such as ``Standing'', are with certain noise. It may be due to the poor quality of the observations of ``Standing'' with noise, where the high-frequency noise existing in observation may affect the generation of the prior normal.

\subsection{Limitations}

We also noticed that the proposed method did not achieve the best performance on some objects of the DiLiGenT dataset \cite{shi2019benchmark}, such as ``Ball'' and ``Cat'', as shown in Fig. \ref{fig8}. We argue that these objects usually have few regions with non-Lambertian reflectance (specularities and shadows). In this case, our method does not outperform others on MAE, such as IRPS \cite{taniai2018neural} and DPSN \cite{santo2017deep}. However, our error map of ``Cat'' shown in Fig. \ref{fig8} is more robust than the others: compared with others, our method handles better in the non-convex regions (crinkles). Also, as shown in Fig. \ref{fig3}, our method with prior normal reports a slightly worse MAE than counterpart without priors, on very few materials in MERL BRDFs dataset \cite{matusik2003data}, which show almost diffused properties (such as ``teflon'', ''pink-fabric'', and ``nylon''). It also illustrates that our method may exist limitations on very slight non-Lambertian samples.

\section{Conclusions and Future Work}

In this paper, we have proposed a Lambertian normal guided photometric stereo network, which utilizes the prior normal to derive accurate surface measurement. Compared with previous deep learning approaches that derive the normal space from the RGB space, our method takes the mapping in the same normal space and pays more attention to the errors in the prior normal. Ablation experiments have illustrated that our method performs more accurate reconstruction. Moreover, the convergence of our method is faster than the traditional methods using the observations to derive the surface normal, which means that our method can be trained with fewer samples. Extensive quantitative and qualitative comparisons on both real (the DiLiGenT benchmark and the Light Stage Data Gallery) and synthetic datasets (the ``Dragon'' and the CyclesPSTest) have shown that our method outperforms the state-of-the-art methods. The examples have demonstrated that our Lambertian priors photometric stereo network better handles the surface normal in strong non-Lambertian materials and surfaces with varying reflectance. In future work, we will explore several alternative normals as the priors. For example, better priors further improve the estimation of the surface normals, such as the high-quality photometric stereo with outlier rejection.  

Furthermore, the proposed priors guided photometric stereo network can also support wider applications. For instance, our framework can be used in a non-ideal illumination environment, where the illuminations are not parallel light or with extra natural illumination. In these tasks, the priors normal can be calculated under the ideal illumination assumption and then refined with the network.

\bibliographystyle{IEEEbib}
\bibliography{bibinfo}

\end{document}